\def\year{2020}\relax
\documentclass[letterpaper]{article} 
\usepackage{aaai20}  
\usepackage{times}  
\usepackage{helvet} 
\usepackage{courier}  
\usepackage[hyphens]{url}  
\usepackage{graphicx} 
\usepackage{graphicx}  
\usepackage{multirow}
\usepackage{amsmath}
\usepackage{amssymb}
\usepackage{pifont}
\usepackage{todonotes} 
\usepackage{multirow}
\newcommand{\x}{\boldsymbol{x}}
\newcommand{\y}{\boldsymbol{y}}

\newcommand{\z}{\boldsymbol{z}}
\newcommand{\w}{\boldsymbol{w}}
\newcommand{\cb}{\boldsymbol{c}}

\newcommand{\cmark}{\ding{51}}%
\newcommand{\xmark}{\ding{55}}
\title{Learning Graph-Based Priors for Generalized Zero-Shot Learning}

\author{Colin Samplawski,\textsuperscript{\rm 1,2} Jannik Wolff,\textsuperscript{\rm 1,3} Tassilo Klein\textsuperscript{\rm 1}, Moin Nabi\textsuperscript{\rm 1}\\
\textsuperscript{\rm 1}SAP ML Research Berlin, \textsuperscript{\rm 2}University of Massachusetts Amherst, \textsuperscript{\rm 3}TU Berlin \\ 
csamplawski@cs.umass.edu,  jannik.wolff@campus.tu-berlin.de, \{tassilo.klein,  m.nabi\}@sap.com
}

\begin{document}
\maketitle
\begin{abstract}
The task of  zero-shot learning (ZSL) requires correctly predicting the label of samples from classes which were unseen at training time. This is achieved by leveraging side information about class labels, such as label attributes or word embeddings. Recently, attention has shifted to the more realistic  task of generalized ZSL (GZSL) where test sets consist of seen and unseen samples. Recent approaches to GZSL have shown the value of generative models, which are used to generate samples from unseen classes. In this work, we incorporate an additional source of side information in the form of a relation graph over labels. We leverage this graph in order to learn a set of prior distributions, which encourage an aligned variational autoencoder (VAE) model to learn embeddings which respect the graph structure. Using this approach we are able to achieve improved performance on the CUB and SUN benchmarks over a strong baseline.
\end{abstract}

\begin{figure}[h!]
\centering
\includegraphics[width=\columnwidth]{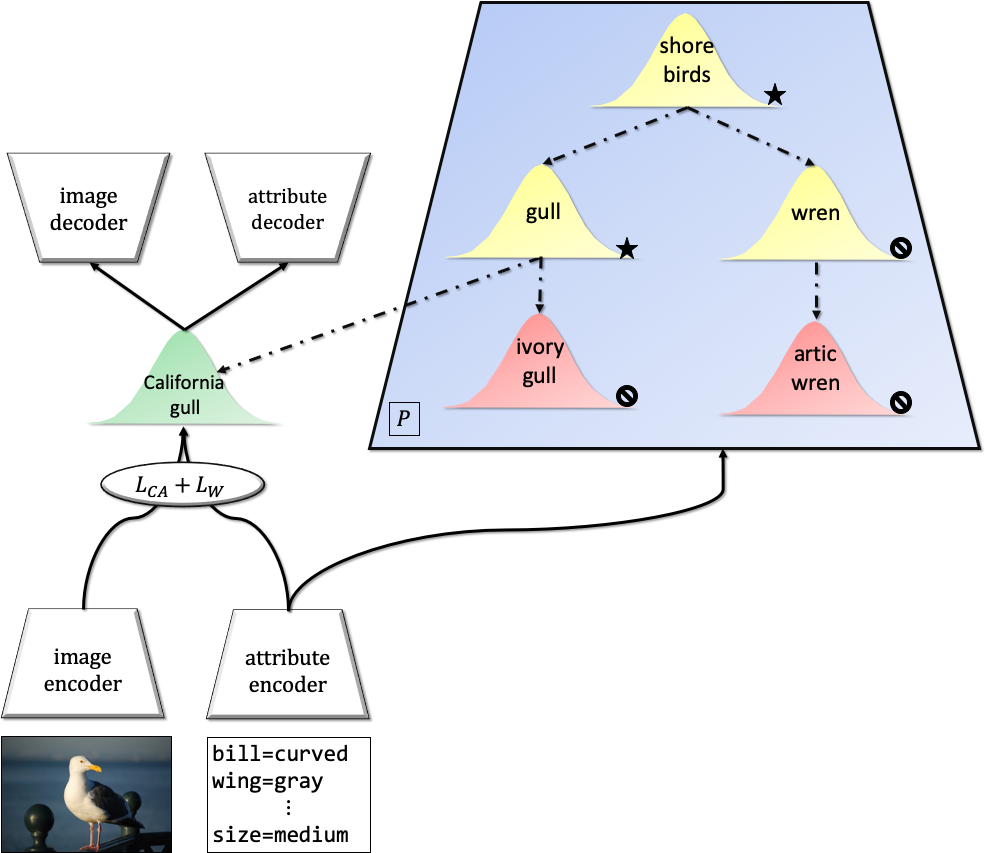}
\caption{A diagram of our method using a simplified graph for the CUB dataset. Each node in the graph is embedded as a Gaussian distribution, with dashed lines representing ``is-a'' relations. The green California gull class has images at training time (i.e. it is in $\mathcal{Y}^S$). The red nodes have no images at training time, but do at test time ($\mathcal{Y}^U$). The yellow nodes are superclasses which have no associated images at any time ($\mathcal{Y}^I$). The blue shaded region is the set of distributions $P$ which make up the prior. For this example, nodes marked with $\star$ are positive samples for California gull and nodes marked with $\O$ are negative samples.}
\end{figure}

\section{Introduction}
Recent work in computer vision has utilized deep learning to achieve better than human performance on many tasks. However, deep models still require a large amount of labeled training data in order to achieve such performance. Methods which can alleviate this requirement therefore represent an active area of research.

The most extreme example of such a setting is zero-shot learning (ZSL). In ZSL, we are interested in correctly making predictions for samples from classes which were unseen at training time. This is achieved by leveraging side information about classes, usually in the form of label attribute vectors or word embeddings. 
In traditional ZSL, test time predictions are made only amongst the unseen classes. However, a more realistic setting is generalized ZSL (GZSL) where test sets include samples from both the seen and unseen classes.

The performance of a zero-shot model is naturally very dependent upon on the quality of side information used to train it \cite{akata2015evaluation}. Previous approaches have shown that leveraging additional side information in the form of a graph over labels can lead to significant gains in zero-shot learning performance \cite{Wang2018,kampffmeyer2019rethinking}.

Orthogonally, the dominant paradigm in recent approaches to GZSL rely on generative models. By conditioning a generative model (such as a generative adversarial network or a variational autoencoder (VAE)) on the label attributes, synthetic samples from unseen classes can be generated to help compensate for the missing data.

To the best of our knowledge, these two ideas have not yet been unified. In this work we offer a straightforward approach which leverages the graph structure in a generative model, namely a VAE-based one. We learn a Gaussian distribution for each node in the graph to act a set of priors for the VAE models. We simultaneously optimize a loss which encourages the node embeddings to respect the graph structure while we train the VAE models. This leads to a shared latent space which respects the graph structure. Our experiments show that this approach leads to improved performance on several GZSL benchmarks over a strong baseline.

\begin{table*}[h!]
	\centering
	\begin{tabular}{|c|c|c|c|c|c|c|} 
		\hline
		Dataset & $|\boldsymbol{a}_c|$ & $|\mathcal{Y}^S|$ & $|\mathcal{Y}^U|$ & $|G|$ & $|\mathcal{D}_{tr}|$ & $|\mathcal{D}_{ts}|$ \\ \hline
		CUB & 312 & 150 & 50 & 200+182 & 7057 & 2967 + 1764\\
		SUN & 102 & 645 & 72 & 717+16 &10320 & 1440 + 2580\\ \hline
	\end{tabular}
	\caption{Dataset statistics. $|\boldsymbol{a}_c|$ denotes the dimensionality of the attribute vectors for each dataset. $|G|$ denotes the number of nodes which have images plus the number of nodes which do not. $|\mathcal{D}_{tr}|$ denotes the number of training instances. Similarly, $|\mathcal{D}_{ts}|$ gives the number of unseen test samples plus the number of seen test samples.}
\end{table*}

\section{Background}
\subsection{Generalized Zero-Shot Learning}
In the standard GZSL problem we have a set of seen classes, $\mathcal{Y}^S$, and unseen classes,  $\mathcal{Y}^U$. For all classes $\mathcal{Y}^S \cup \mathcal{Y}^U$ we have access to side information usually in the form of label attribute vectors, $\mathcal{A} = \{\boldsymbol{a}_c \ | \ c \in \mathcal{Y}^S \cup \mathcal{Y}^U \}$. We further have access to a dataset $\mathcal{D} = \{(\x,y) \ | \ \x \in \mathcal{X}, y \in \mathcal{Y}^S\}$ of training samples from the seen classes, where $\mathcal{X}$ is the image feature space (such as the final hidden layer of a deep CNN model). Using $\mathcal{D}$ and $\mathcal{A}$, the goal of the learner is then to learn a classification function $f : \mathcal{X} \rightarrow \mathcal{Y}^S \cup \mathcal{Y}^U$.

Early approaches to ZSL sought to learn an inductive compatibility function between image features and label attributes \cite{Frome2013,Norouzi2013,Socher2013}. Then, at test time, the compatibility score between a test image and all possible labels is calculated. The most compatible label is then predicted.

\subsection{Graphical Approaches to GZSL}
We can expand the traditional GZSL definition by adding an additional source of side information in the form a directed relation graph $G$. This graph has nodes corresponding to $\mathcal{Y}^S$ and $\mathcal{Y}^U$ together with an additional set of nodes $\mathcal{Y}^I$ for which no images exist (neither at training nor test time). Nodes in $G$ further have at least an associated attribute vector. Finally, an edge in $G$ indicates that there is a relation between two nodes. A simple example of such a graph is a tree-structured label hierarchy where edges indicate ``is-a'' relations between classes, and $\mathcal{Y}^I$ is the set of superclasses of $\mathcal{Y}^S \cup \mathcal{Y}^U$, which are assumed to be leaves. We note however that our approach does not require a hierarchy, but rather can be applied to any DAG over labels. 

Some previous approaches to ZSL have recognized the benefit of leveraging such a graph as another source of side information. In the simplest form, this can be a single vector, which encodes a node's adjacency information to all other nodes \cite{akata2015evaluation}. More recently, several approaches have used graph convolution networks to predict classifier weights for unseen classes by using the label attributes as input \cite{Wang2018,kampffmeyer2019rethinking}.
A limitation of these approaches compared to ours is that since their representation of the graph is non-probabilistic, they are unable to encode uncertainty about relations in the graph. Furthermore, none of these models are generative approaches. 

\subsection{Generative Approaches to GZSL}
Most recently, generative approaches to GZSL have shown impressive improvements in performance over  earlier compatibility-based approaches. Using a generative model conditioned on the label attribute information, one can generate samples for the classes, which were unseen at training time. Afterwards, a classifier can be trained using real and generated samples. This has been successfully applied using generative adversarial networks \cite{Xian_2018_CVPR,sariyildiz2019gradient} and variational autoencoders \cite{mishra2018generative,cadavae}. Most relevant to our work is the cross-alignment and Distribution-Alignment VAE (CADA-VAE) model \cite{cadavae}, which our approach builds upon (Section 3).

\subsubsection{Variational Autoencoders}
A variational autoencoder \cite{Kingma2014} aims to learn the conditional distribution $p_{\theta}(\z | \x)$ for each data sample $\x$ and latent factors $\z$. Since the latent factors are generally not known, we estimate this distribution via the parameterized distribution $q_{\phi}(\z | \x)$. This is accomplished by maximizing the evidence lower-bound (ELBO):
\begin{equation}
\mathbb{E}_{q_{\phi}(\z | \x) } \left[ \log p_{\theta} (\x | \z) \right]
- D_{KL} (q_{\phi}(\z | \x) || p_{\theta}(\z))
\end{equation}
The first term is the loss of reconstructing $\x$ from the predicted latent factors $\z$ using the decoder $ p_{\theta} (\x | \z)$. The second term minimizes the KL divergence between the encoder $q_{\phi}(\z | \x)$ and a prior $p_{\theta}(\z)$, which acts as a regularizer. 

\subsubsection{CADA-VAE}
The CADA-VAE model consists of a VAE for each modality of the data (i.e. images and attributes). Consider the case for a pair of modalities $i$ and $j$. Let $Q^i$ and $D^i$ be the encoder and decoder for modality $i$, respectively. They use L1 reconstruction loss as a surrogate for $\log p_{\theta}(\x | \z)$ and the standard normal distribution as a prior. So, for each VAE we have the following loss:
\begin{align}
\begin{split}
L_{VAE}^i &=  |\x^i - D^i(Q^i(\x^i))| \\
&- \alpha D_{KL}(Q^i(\x^i) || \mathcal{N}(0, I))
\end{split}
\end{align}
where $\x^i$ is a training sample from modality $i$.
They then seek to align the latent spaces of the VAEs using a cross-alignment loss where each decoder must successively decode the generated latent samples for each other modality. The cross-alignment loss is then:
\begin{equation}
L_{CA} = |\x^j - D^j(Q^i(\x^i))| +  |\x^i - D^i(Q^j(\x^j))|
\end{equation}
where $\x^i$ and $\x^j$ are samples from each modality with the same class.
They further enforce alignment by adding a Wasserstein distance penalty between each pair of encoded distributions from different modalities but with the same class. 
\begin{equation}
L_{W} = W_2(Q^i(\x^i), Q^j(\x^j))
\end{equation}
Since both distributions are multivariate Gaussians, this can be computed efficiently in closed form.
 The full CADA-VAE loss is then:
\begin{align}
L_{CADAVAE} = L_{VAE}^i  + L_{VAE}^j  + \beta L_{CA} + \gamma L_{W}
\end{align}

\subsection{Gaussian Embeddings}
Given an relation graph is it often useful to learn a representation for each node. This is commonly achieved by minimizing some distance between nodes which have a relation (the positive samples) and maximizing the distance between nodes which do not have a relation (the negative samples) \cite{paccanaro2001learning}. For example, it has been shown that this can be effectively done when embedding each concept as a multivariate Gaussian distribution \cite{Vilnis2015} . They optimize an energy based max-margin ranking objective for margin $m$ given by:
\begin{equation}
L_m(\w, \cb_p, \cb_n) = \max(0, m - E(\w, \cb_p) + E(\w, \cb_n))
\end{equation}
Here $\w$ is the embeddings of the main ``word'' and $\cb_p$ and $\cb_n$ are positive and negative context words, respectively. In the case of the relation graphs, $\w$ corresponds to a node in the graph and $\cb_p$ / $\cb_n$ are embeddings of other nodes which do / do not have a relation with $\w$. The energy function $E$ computes how similar a pair of embeddings are. Relations are often asymmetric, so it is often sensible to use an asymmetric metric for the energy function \cite{Vilnis2015}, namely the negative Kullback-Leibler divergence:
\begin{equation}
E(\x, \y) = -D_{KL}(\x || \y)
\end{equation}

Such embeddings have shown success on entailment \cite{Vilnis2015}, link prediction, and node classification \cite{Bojchevski2018} tasks. 
\section{Our Approach}
Our approach is then to unify the aligned VAEs of CADA-VAE together with the Gaussian embeddings approach (Figure 1). Rather than using single global prior when training the VAEs, we associate a Gaussian distribution for each node in $G$ which has no images available at training time to act as priors for the seen classes. We then use a similar ranking objective to push the seen class distributions of both modalities to a location which respects the graph structure as encoded by the priors. Let $P = \{\mathcal{N}_c \ | \ c \notin \mathcal{Y}^S \}$ be the set of Gaussian distributions that make up the prior. We replace the VAE loss  with the following term:
\begin{align}
\begin{split}
\hat{L}_{VAE}^i &=  |\x^i - D^i(Q^i(\x^i))| \\
 &+ \alpha [ L_m(Q^i(\x^i) , \cb_p, \cb_n) ) \\
 &-D_{KL}(Q^i(\x^i) || \mathcal{N}(0,I))]
 \end{split}
\end{align}
where $\cb_p \in P$ is a Gaussian distribution for a node which has a relation with the class of $\x^i$. Likewise $\cb_n \in P$ is a Gaussian distribution for a node which does not a relation. We still include the typical $\mathcal{N}(0,I)$ prior, but here it can be thought of as  a kind of root node in the graph with which all nodes have a relation.

We have not yet specified how one can obtain the set $P$. One simple approach would be to pretrain these embeddings ahead of time and keep them fixed while training the VAEs. However, we suggest an end-to-end approach where the distributions making up the prior are learned together with the VAE. In fact, we predict the parameters for the distributions in $P$ using the attribute encoder. That is, for each $\cb_i \in P$, we have $\cb_i = Q^{\mathcal{A}}(\boldsymbol{a}_{i})$. We then add an additional loss to our objective which encourages the distributions in $P$ to encode the graph structure. For a distribution $\cb_i \in P$, we have:
\begin{align}
L_{prior}= L_m(\cb_i, \cb_p, \cb_n) -D_{KL}(\cb_i || \mathcal{N}(0,I)) 
\end{align}
where again $\cb_p$ / $\cb_n$ are nodes with positive / negative relations with $\cb_i$. So our total objective is then:
\begin{equation}
L_{ours} = \hat{L}_{VAE}^i  + \hat{L}_{VAE}^j  + \beta L_{CA} + \gamma L_{W} + \epsilon L_{prior}
\end{equation}

\begin{table*}[t!]
	\centering
	\begin{tabular}{|c|c|ccc|ccc|} 
		\hline
		\multirow{2}{*}{Model} & \multirow{2}{*}{Graph} & \multicolumn{3}{c|}{CUB}  & \multicolumn{3}{c|}{SUN} \\
		& & U & S & H & U & S & H \\ \hline
		CADA-VAE & - & 51.6 & 53.5 &  52.4 & 47.2 & 35.7 & 40.6  \\
		Ours & Flat & 47.8 & 59.0 &  52.8 & 44.0  & 38.3 &  40.9  \\
		Ours & Full & 51.4 & 58.3 &  \textbf{54.6} & 45.0 & 38.0 & \textbf{41.2}  \\ \hline
	\end{tabular}
	\caption{The results of our experiments. U and S denote the average top-1 per class accuracy for the seen and unseen classes, respectively. H is the harmonic mean between these two values. The results for CADA-VAE are copied from the original paper.}
\end{table*}
\section{Experiments}
We present results using two common GZSL benchmarks, the Caltech-UCSD Birds dataset (CUB) \cite{cub} and the SUN with attributes dataset (SUN) \cite{patterson2012sun} (Table 1). We use the pretrained ResNet101 features provided by \cite{Xian2017}. For CUB, we use the class hierarchy provided by \cite{cubhierarchy} which provides a tree of ``is-a'' relations over the species in CUB which is derived from the Wikispecies dataset. The SUN dataset includes a ``is-a'' graph over classes, however, it is a not a hierarchy, with some nodes having multiple parents. Both graphs contain nodes which have no associated attributes. We set these nodes' attributes to be the mean of the attributes of all nodes which are reachable from that node. 

Aside from including the graph structure, our approach adds one additional change compared to CADA-VAE. Namely, the attribute encoder is used to encode attributes from unseen classes during training. This will likely improve performance on its own, so we present results using two graphs. We first use a flat, uninformative graph where all nodes have a positive relation only with a root node. The ranking objective then only encourages the distributions for each class to be far away from each with respect to the KL divergence. This leads to a model which has some similarity to the one in \cite{wang2018zero}.  We further provide results using the full graph for each dataset.

\subsection{Implementation Details}
We chose to stick close to the best performing hyperparameters reported for CADA-VAE. The encoders predict a 64-dimensional multivariate Gaussian with  diagonal covariance. 
The image encoder was a one layer MLP with 1560 hidden units. The image decoder was a one layer MLP with 1660 hidden units. Similarly, the attribute encoder was a one layer MLP with 1450 units and the attribute decoder has 660 hidden units. The VAE models were trained for 100 epochs using Adam with a learning rate of 0.00015 and a batch size of 50. The loss weight scales were set with an annealing scheme where all scaled losses started with 0 weight and are then increased linearly for a set number of epochs. The scale on the KL divergence term ($\alpha$)  increased by 0.003 for each epoch in (0, 93]. The scale on the cross-alignment loss ($\beta$) was increased by 0.045 for each epoch in (6, 75], and the weight scale on the Wassterstein distance term ($\gamma$) was increased by 0.55 for each epoch in (6, 22]. These hyperparameters were kept fixed and the only new hyperparameters we introduced were similar weighting parameters for the prior loss ($\epsilon$). These values were tuned separately for each dataset via random search using the provided validation split. On both datasets $\epsilon$ is increased in epochs $(0,49]$. The value is increased by 0.0717 for CUB and 0.02271 on SUN. The margin $m$ in the $L_m$ loss was set to 1. 

The final classifier was a logistic regression model  also trained with Adam for 20 epochs with a learning rate of 1e-3 and batch size of 32. The dataset for the classifier was generated using the encoders of the VAE models. For the seen classes, 200 samples were drawn for each class using random training images as input to the encoder. For the unseen classes 400 samples were drawn from each distribution predicted for each attribute vector. 

The positive and negative context pairs for each node were generated using the full transitive closure of the graph. That is, two nodes have a relation if there exists a path between those two nodes (Figure 1). At each training step, in addition the batch of image / attribute pairs from seen classes, we sample a batch of the same size from nodes in $P$ in order to optimize $L_{prior}$. 

\begin{table}[t!]
	\centering
	\begin{tabular}{|c|cc|ccc|} 
		\hline
		\multirow{2}{*}{Model} & \multirow{2}{*}{$L_{CA}$} & \multirow{2}{*}{$L_{DA}$}& \multicolumn{3}{c|}{CUB} \\
		& &  & U & S & H \\ \hline
		\multirow{4}{*}{CADA-VAE} & \xmark & \xmark & 0.13  & 67.1 & 0.27 \\
		 & \cmark & \xmark & 48.1 & 52.6 &  50.2 \\
		 & \xmark & \cmark & 43.8 & 48.1 &  45.8 \\
		 & \cmark & \cmark & 51.6 & 53.5 &  52.4 \\ \hline
		\multirow{4}{*}{Ours} & \xmark & \xmark & 9.88 & 60.8 & 17.0\\
		 & \cmark & \xmark & 13.6 & 69.3 & 22.8  \\ 
		 & \xmark & \cmark & 39.3 & 40.8 & 40.1  \\ 
		 & \cmark & \cmark & 51.4 & 58.3 &  \textbf{54.6}  \\ \hline
	\end{tabular}
	\caption{Ablation study of loss terms on the CUB dataset.}
\end{table}
\section{Results and Discussion}
We first present qualitative results in the form of a 2D TSNE projection \cite{maaten2008visualizing} of the learned latent space on the CUB dataset as shown in Figure 2. In particular, we plot the projection of the mean of the predicted distribution for each attribute. We focus on the latent space made up of three subgroups which fall under the ``Charadriiformes'' superclass, also known as ``shore birds''. We see that the model has learned to group the fine-grained leaf classes together with the embedding of the supercategory placed roughly equidistant from them. We see a similar phenomenon at a higher semantic level where the ancestor of all these nodes is situated roughly in the middle of the three categories.

The main quantitative results of our experiments are shown in Table 2. We see that using our approach with only a simple flat graph does result in small gains in performance over CADA-VAE, most likely to do the fact that the unseen class attributes are used to help train the attribute encoder. However, we see a further increase in performance when using the full graph on both datasets. We see the strongest gains on the CUB dataset. This may be due to the fact that the CUB graph is a tree, whereas the graph over SUN is a DAG, which is likely more difficult to encode. Furthermore, the graph provided for CUB is much denser, with 182 internal nodes defining the supercategory structure of the labels. By contrast, the SUN graph adds only 16 additional higher level nodes.

Finally, we include an ablation study over the alignment loss functions our approach uses (Table 3). In particular, we compare against CADA-VAE in three scenarios using the CUB dataset. First we consider removing the $L_{CA}$ loss from the objective, removing the $L_{DA}$ loss, and finally removing both. We see that CADA-VAE is more robust to turning off loss terms compared to our approach. In particular, the success of our prior seems very dependent on also including the cross-alignment loss.  When we turn off both alignment losses, the CADA-VAE is unable to maintain performance, achieving a nearly zero harmonic mean. This is unsurprising, since without enforcing any alignment at all, it is nearly impossible to generalize to the unseen classes. However, under the same setting our approach is still able to maintain some performance with respect to harmonic mean. This suggests that the prior itself can provide an implicit alignment of the two VAEs by encouraging both of them to respect the graph structure in the same way.


\begin{figure}
\centering
\includegraphics[width=.85\columnwidth]{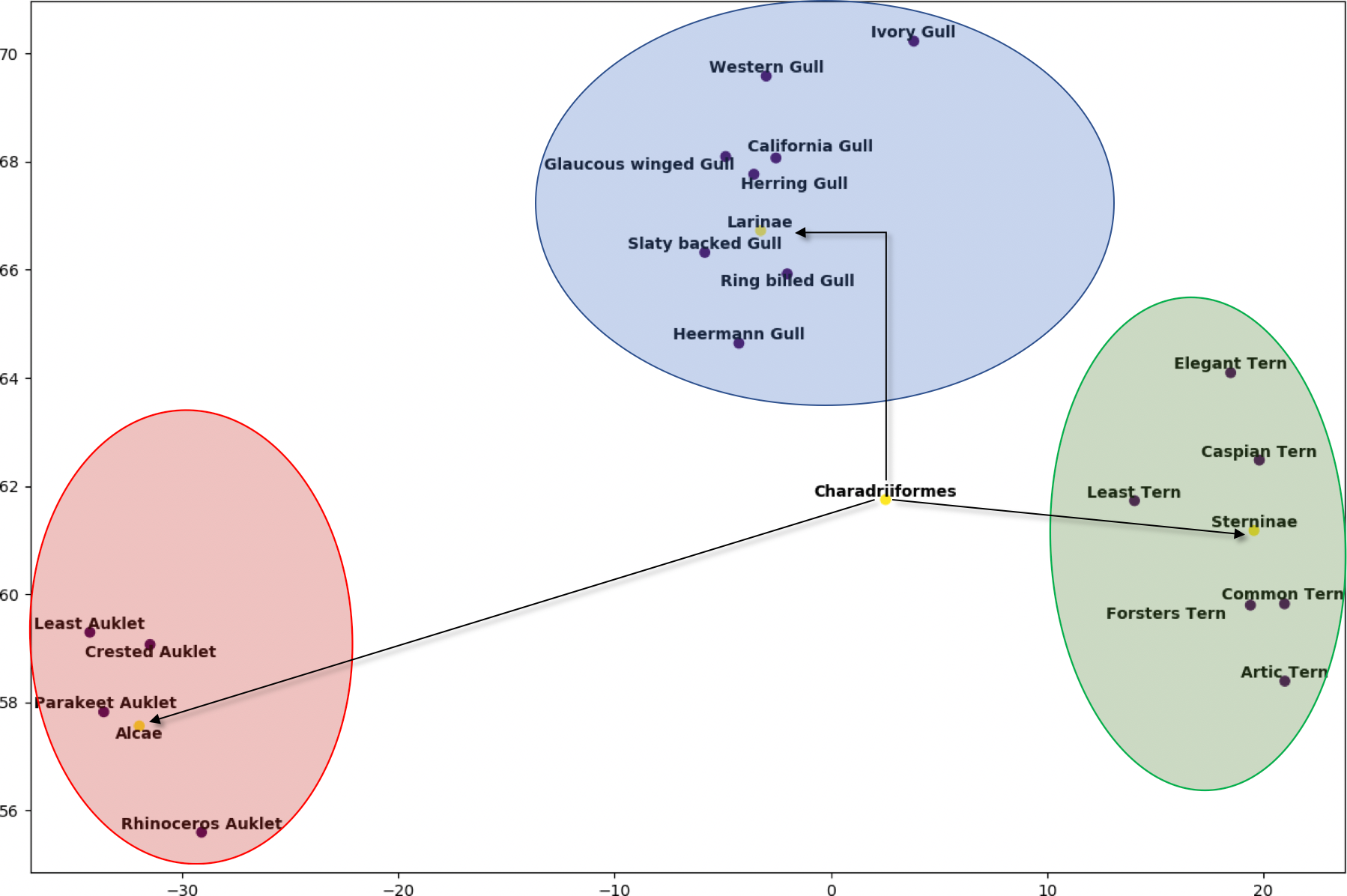}
\caption{TSNE projection of the learned means from the attribute encoder. The dark purple points indicate leaf nodes. The light yellow nodes indicate superclasses. We display a simplified subgraph of the full CUB graph rooted at the ``Charadriiformes'' (shore birds) node. It is split into three subcategories: gulls, wrens, and auklets, as indicated by the colored ovals.}
\end{figure}

\section{Conclusion}
We have provided a straightforward way to incorporate the structure of a graph over labels in a generative model for GZSL. We do so by learning a Gaussian embedding for each node in the graph to act as a prior for an aligned VAE model. We show that by training the distributions that make up this prior end-to-end with the VAE models, we are able to achieve improved performance on several GZSL benchmarks over a strong baseline.
\bibliography{main}
\bibliographystyle{aaai}
\end{document}